\documentclass[review]{elsarticle}

\usepackage{lineno,hyperref}
\usepackage{url}            
\usepackage{booktabs}       
\usepackage{amsfonts}       
\usepackage{nicefrac}       
\usepackage{microtype}      
\usepackage{amsmath}
\usepackage{multirow}
\usepackage{amssymb}
\usepackage{graphicx}
\usepackage{wrapfig}
\usepackage{soul}
\usepackage[dvipsnames]{xcolor}

\usepackage{makecell}
\usepackage{placeins}
\usepackage{caption} 

\modulolinenumbers[5]

\newtheorem{lemma}{Lemma}
\newtheorem{theorem}{Theorem}[section]

\newtheorem{Algorithm}{Algorithm}[section] 
 
\newtheorem{Remarks}{Remark}[section]

\journal{Information Sciences}

\begin{document}

\begin{frontmatter}

\title{DPER: Efficient Parameter Estimation for Randomly Missing Data}


\author[firstaddress]{Thu Nguyen}
\author[seventhaddress,eighthaddress,ninthaddress]{Khoi Minh Nguyen-Duy}
\author[secondaddress]{Duy Ho Minh Nguyen}
\author[seventhaddress,eighthaddress,ninthaddress]{Binh T. Nguyen\corref{correspondingauthor}}
\cortext[correspondingauthor]{Two first authors have equal contribution. The corresponding author: Binh T. Nguyen}
\ead{ngtbinh@hcmus.edu.vn}
\author[firstaddress]{Bruce Alan Wade}
\address[firstaddress]{University of Louisiana at Lafayette}
\address[secondaddress]{Max Planck Institute for Informatics, Germany}
\address[seventhaddress]{AISIA Research Lab, Ho Chi Minh City, Vietnam}
\address[eighthaddress]{University of Science, Ho Chi Minh City, Vietnam}
\address[ninthaddress]{Vietnam National University, Ho Chi Minh City, Vietnam}



%
%

\begin{abstract}
The missing data problem has been broadly studied in the last few decades and has various applications in different areas such as statistics or bioinformatics. Even though many methods have been developed to tackle this challenge, most of those are imputation techniques that require multiple iterations through the data before yielding convergence. In addition, such approaches may introduce extra biases and noises to the estimated parameters. In this work, we propose novel algorithms to find the maximum likelihood estimates (MLEs) for a one-class/multiple-class randomly missing data set under some mild assumptions.    

As the computation is direct without any imputation, our algorithms do not require multiple iterations through the data, thus promising to be less time-consuming than other methods while maintaining superior estimation performance. We validate these claims by empirical results on various data sets of different sizes and release all codes in a GitHub repository to contribute to the research community related to this problem.

\end{abstract}


\end{frontmatter}

\section{Introduction}
\label{sec:introduction}

Missing data is a common problem in data analysis. For instance, respondents in a survey may refuse to fill in the income, or sensors may fail and need to be replaced. These imply missing data challenges.  
Most of the existing methods \cite{marivate2007autoencoder, buuren2010mice, kim2005missing, mazumder2010spectral, candes2009exact, rahman2016missing, burgette2010multiple, costa2018missing} require revisiting the data multiple times before yielding convergence and are computationally expensive in high-dimensional settings.

In this paper, we introduce a novel approach, namely DPER (\textit{Direct Parameter Estimation for Randomly Missing Data}), that can compute the maximum likelihood estimators (MLEs) for both mean and covariance matrices of a one-class/multiple-class data set. For a multiple-class data set, we develop solutions for both cases when all classes' covariance matrices are assumed to be equal or not. Since the computation is direct, our approach does not require many iterations through the data, promising a faster estimation speed. For the same reason, unlike many other imputations methods that may run slower when there are more missing entries, our approach is more rapid when there are more missing entries. Yet, experiments show that the resulting estimates are far superior to many state-of-the-art methods. In addition, we do not assume multivariate normality as presented in \cite{NGUYEN20211}. Instead, we use a milder assumption of bivariate normality in each pair of features. Lastly, it is worth noting that normal distribution is a widely used approximator due to the Central Limit Theorem when the data's distribution is unknown. Therefore, even under the fore-mentioned mild assumption, our approach is robust enough to be applied in various applications, as illustrated in our experiments later.

Finally, the contributions of our work can be summarized as follows:
\begin{enumerate}[(i)]
\item We compute the MLEs directly for the mean and covariance matrix of one-class/multiple-class randomly missing data. Therefore, we provide an approximation to the underlying distribution of the data.
\item We provide algorithms to compute the MLEs from the data that are suitable for both small and large data sets.
\item The experimental results show that our algorithms can outperform state-of-the-art methods in error rates, speed, and memory requirements.
\item Our experiments on large data sets show that while iterative methods may not handle the missing data problem for big data sets, our approach can still find the parameter rapidly without requiring extensive memory and computation time.
We discuss the potential drawbacks of our methods and future research directions.
\end{enumerate}

The rest of this paper can be organized as follows. Section \ref{sec:related_work} introduces an overview of the current researches related to the problem. In Section \ref{sec: without}, we present the formulas and algorithms for computing the means and covariance matrix for a randomly missing data set that only consists of a single class and when there are multiple classes without assuming the equal covariance matrices are equal. Next, we present the estimation of the parameters under the assumption of equal covariance matrices in a multiple-class setting. Finally, we illustrate the power of our approach via experiments in Section \ref{exper} and discuss the implications of the experimental results. The paper ends with our conclusion and future works in Section \ref{sec:conclusion}.

\section{Related works}
\label{sec:related_work}
Various methods have been developed to tackle the problem of missing data. They can be divided into several subcategories. The first category includes all techniques that are based on matrix completion such as Polynomial Matrix Completion \cite{fan2020polynomial}, SOFT-IMPUTE \citep{mazumder2010spectral}, and Nuclear Norm Minimization \citep{candes2009exact}. 
The second category consists of regression-based and least-square techniques, such as, CBRL and CBRC \cite{m2020cbrl}, 
multiple imputation by chained equation (MICE)\citep{buuren2010mice}, local least squares imputation \citep{kim2005missing}, and least trimmed squares imputation \citep{templ2011iterative}. 
The third one consists of tree-based techniques such as sequential regression trees \citep{burgette2010multiple}, missForest \cite{stekhoven2012missforest}, DMI algorithm \cite{rahman2013missing}, and C4.5 algorithm \citep{fortes2006inductive} that are applied for this problem. Next, the fourth category contains clustering-based techniques such as Fuzzy c-means clustering \citep{aydilek2013hybrid}, K-nearest neighbor imputation (KNN)\citep{garcia2010pattern}, kEMI and $\text{kEMI}^+$ \citep{razavi2020similarity}, imputation using fuzzy neighborhood density-based clustering \citep{razavi2016imputation}, imputation using Hybrid K-Means and Association Rules\citep{chhabra2018missing}, and evolutionary clustering method \cite{gautam2015data}. 
Also, one can find techniques that are based on extreme learning, for example, AAELM \cite{gautam2015data}, extreme learning machine multiple imputation \cite{sovilj2016extreme}, or Bayesian approaches, which are Naive Bayesian imputation \citep{garcia2005naive}, Bayesian network imputation \citep{hruschka2007bayesian}, and Bayesian principal component analysis-based imputation\citep{audigier2016multiple}.
 
In addition, deep learning techniques also form another dominant class of imputation methods. Those techniques include multiple imputations using Deep Denoising Autoencoders \citep{gondara2017multiple}, imputation via Stacked Denoising Autoencoders \citep{costa2018missing}, imputation via Adversarially-trained Graph Convolutional Networks \citep{spinelli2020missing}, a Swarm Intelligence-deep neural network \citep{leke2016missing}, combining Gravitational search algorithm with a deep-autoencoder \citep{garg2018dl}, Generative Adversarial Multiple Imputation Network \cite{yoon2020gamin}, etc. Hence, they are not applicable to small data sets.

Last but not least, the final category consists of fuzzy imputation techniques
 such as the evolving granular fuzzy-rule-based model 
\cite{garcia2019evolvable}, \cite{garcia2019incremental}, fuzzy c-means clustering \citep{aydilek2013hybrid},
fuzzy clustering-based EM imputation\citep{rahman2016missing}, decision trees and fuzzy clustering with iterative learning (DIFC) \cite{nikfalazar2020missing},
and imputation using fuzzy neighborhood density-based clustering \citep{razavi2016imputation}.

Up to now, there have been several attempts to estimate the parameters directly from the data \citep{anderson1957maximum,fujisawa1995note, NGUYEN20211}. Yet, to our knowledge, these approaches rely on the assumption that the data follow a multivariate normal distribution and can only be used for monotone missing data. On the other hand, our works only assume pairwise normality (i.e., each pair of features follows a bivariate normal distribution) and work for randomly missing data in general.

For the rest of this paper, $\hat{\theta}$ denotes the maximum likelihood estimate of a parameter $\theta$, $A$' denotes the transpose of a matrix $A$, $B^{-1}$ indicates the inverse of a square matrix $B$. 

\section{Methodology}\label{sec: without}

This section will describe our proposed approaches to find the maximum likelihood estimators for a randomly missing data set that only consists of a single class and for a multiple-class data set without the assumption of equal covariance matrices. 

\subsection{Single-class DPER algorithm}\label{sec:1class-method}

First, we have the following results:

\begin{theorem}\label{bivariatemle_single}
	Assume that we have a set of i.i.d observations from a bivariate normal distribution with mean  
	\begin{equation}
		\boldsymbol {\mu } = \begin{pmatrix}
			\mu_1\\ \mu_2
		\end{pmatrix}
	\end{equation}
	and covariance matrix 
	\begin{equation}
		\boldsymbol {\Sigma } =  \begin{pmatrix}
			\sigma_{11}&\sigma_{12}\\
			\sigma _{21} & \sigma _{22}
		\end{pmatrix}.
	\end{equation}	
	and arrange the data into the following pattern
	\begin{equation}\label{blform}
		\boldsymbol{x} = \begin{pmatrix}
			{x}_{11} & ... & {x}_{1m}&{x}_{1m+1}&...&{x}_{1n}&*&...&*\\
			{x}_{21}&...&{x}_{2m}& *& ...& *& {x}_{2n+1}&...&{x}_{2l}
		\end{pmatrix}.
	\end{equation}
	Note that each column represents an observation, and $x_{ij} \in \mathbb{R}$ is an entry, i.e., each observation has two features.  
	
	Let $L$ be the likelihood of the data and	
	\begin{align}
		s_{11} &= \sum_{j=1}^{m}(x_{1j}-\hat{\mu}_1)^2	,\\
		s_{12} &= \sum_{j=1}^{m}(x_{2j}-\hat{\mu}_2)(x_{1j}-\hat{\mu}_1),\\
		s_{22} &= \sum_{j=1}^{m}(x_{2j}-\hat{\mu}_2)^2.
	\end{align}
	
	Then, the resulting estimators obtained by maximizing $L$ w.r.t $\mu_1, \sigma_{11},\mu_2, \sigma_{22}$, and $\sigma_{12}$ are:
	\begin{align*}
		\hat{\mu}_1 & = \frac{1}{n}\sum_ {j=1}^{n}x_{1j},\;\;
		\hat{\mu}_2 = \frac{\sum_ {j=1}^{m}x_{2j}+ \sum_ {j=n+1}^{l}x_{2j}}{m+l-n},	 \\
		\hat{\sigma}_{11} & = \frac{\sum_ {j=1}^n(x_{1j}-\hat{\mu}_1)^2}{n},\\
		\hat{\sigma}_{22} & = \frac{\sum_ {j=1}^{m}(x_{2j}-\hat{\mu}_2)^2+ \sum_ {j=n+1}^{l}(x_{2j}-\hat{\mu}_2)^2}{m+l-n},	
	\end{align*}
	
	and $\hat{\sigma} _{12}$, where $\hat{\sigma} _{12}$ is the maximizer of:	
	\begin{equation}\label{etacov-sing}
		\eta = C-\frac{1}{2}m\log \left(\sigma_{22}-\frac{\sigma_{12}^2}{\sigma_{11}}\right)
		- \frac{1}{2} \left( s_{22} - 2\frac{\sigma_{12}}{\sigma_{11}} s_{12}+\frac{\sigma_{12}^2}{\sigma _{11}^2}s_{11}\right)\left(\sigma_{22}-\frac{\sigma_{12}^2}{\sigma_{11}}\right)^{-1}.
	\end{equation}
\end{theorem}

It is important to note that the estimates for $\mu_1, \mu_2,\sigma_{11}, \sigma_{22}$ are the corresponding sample means and uncorrected sample variance (where the denominator is divided by the sample size) of the first/second feature after dropping the missing entries. We have the following theorem related to the existence of $\hat{\sigma}_{12}$.
\begin{theorem}\label{unique-single}
		In Equation (\ref{etacov-sing}), solving
		\begin{equation}\label{der01}
			\frac{d\eta}{d\sigma_{12}} = 0
		\end{equation}
		can be reduced to solving the following third-degree polynomial 
		\begin{equation}\label{eqExist0}
    	\small
    	s_{12} \sigma_{11} \sigma_{22} + {\left( \sum_{g=1}^Gm_g\sigma_{11} \sigma_{22}  -  s_{22}\sigma_{11} - s_{11}\sigma_{22} \right)}\sigma_{12}  + s_{12}\sigma_{12}^2- (\sum_{g=1}^Gm_g)\sigma_{12}^3,
        \end{equation}
		which has at least one real root. In addition, the global maximum is a real solution to that equation, provided that
\begin{equation}
	-\frac{s_{22}\sigma_{11}+s_{11}\sigma_{22}}{2\sqrt{\sigma_{11}\sigma_{22}}} \neq
	s_{12} \neq \frac{s_{22}\sigma_{11}+s_{11}\sigma_{22}}{2\sqrt{\sigma_{11}\sigma_{22}}}.
\end{equation}
\end{theorem}

The proof of Theorems \ref{bivariatemle_single} and  \ref{unique-single} follow directly as corollaries of Theorems \ref{bivariatemle} and \ref{unique-multiple}, the proofs of which are given in \ref{existence} and \ref{proof2mle}, respectively. Based on the above theorem, we have the following algorithm:

\begin{Algorithm}{(DPER algorithm for single-class randomly missing data)} 
	
	\textit{Input:} a data set of $p$ features. 
	
	\textit{Ouput:} $\hat{\boldsymbol {\mu}}$ and $\hat{\boldsymbol {\Sigma}}=(\sigma_{ij})_{i,j=1}^p$.
	
	\begin{enumerate}		
		
		\item Estimate $\boldsymbol{\mu}$: $\hat{\mu}_i$ is the sample mean of all the available entries in the $i^{th}$ feature. 
		
		\item Estimate the diagonal elements of $\boldsymbol {\Sigma}$: $\hat{\sigma}_{ii}$ is the uncorrected sample variance of all the available entries in the $i^{th}$ feature.  
		
		\item  For $1\le i\le p$:
		
		\;\;\; For $1\le j\le i$:
		
		\;\;\;\;\;\; Compute $\hat{\sigma}_{ij}$ based on Equation  (\ref{der01}). If there are two solutions maximizing the function, choose the one closest to the estimate based on case deletion method.
		
	\end{enumerate}
\end{Algorithm}

Recall that in the \textit{case deletion method}, the sample that has one or more missing values is deleted. In this setting, during the estimation of $\hat{\sigma}_{ij}$, we delete any $(i,j)^{th}$ pair that has at least one missing entry. 

\subsection{Multiple-class DPER algorithm without the assumption of equal covariance matrices}\label{sec: multiple-class-DPER-without}

When the covariance matrices of all classes are not assumed to be equal, the mean and covariance matrix for each class can be estimated separately by applying the \textit{DPER algorithm for single class randomly missing data} to the data from each class.

\subsection{Multiple-class DPER algorithm under the assumption of equal covariance matrices}\label{mgmle}

In the previous section, we have described our proposed approaches to find the maximum likelihood estimators for a randomly missing data set that only consists of a single class and a multiple-class data set without the assumption of equal covariance matrices. Yet, in many cases, for a multiple-class data set, it may be more desirable to assume that the covariance matrices are equal. Those cases may happen when the number of samples per class is small or in linear discriminant analysis. In what follows, we tackle the multiple-class problem under the assumption of equal covariance matrices.

\begin{theorem}\label{bivariatemle}
	Assume that there is a data set with $G$ classes, where each sample from class $g^{th} \;( 1\le g\le G)$ follows a bivariate normal distribution with mean  
	\begin{equation}
	\boldsymbol {\mu }^{(g)} = \begin{pmatrix}
		\mu_1^{(g)}\\ \mu_2^{(g)}
	\end{pmatrix},
\end{equation}
and covariance matrix 
\begin{equation}
	\boldsymbol {\Sigma } =  \begin{pmatrix}
		\sigma_{11}&\sigma_{12}\\
		\sigma _{21} & \sigma _{22}
	\end{pmatrix}.
\end{equation}	
We arrange the data from each class into the following pattern
	\begin{equation}\label{blform}
		\boldsymbol{x}^{(g)} = \begin{pmatrix}
			{x}_{11}^{(g)} & ... & {x}_{1m_g}^{(g)}&{x}_{1m_g+1}^{(g)}&...&{x}_{1n_g}^{(g)}&*&...&*\\
			{x}_{21}^{(g)}&...&{x}_{2m_g}^{(g)}& *& ...& *& {x}_{2n_g+1}^{(g)}&...&{x}_{2l_g}^{(g)}
		\end{pmatrix}.
	\end{equation}
	Note that each column represents an observation and $x_{ij} \in \mathbb{R}$, i.e., each observation has only two features.  

Let $L$ be the likelihood of the data and
\begin{align}
	A &= \sum_{g=1}^Gm_g,\\
	s_{11} &= \sum_{g=1}^G\sum_{j=1}^{m_g}(x_{1j}^{(g)}-\hat{\mu}_1^{(g)})^2	
	,\\
	s_{12} &= \sum_{g=1}^G\sum_{j=1}^{m_g}(x^{(g)}_{2j}-\hat{\mu}_2^{(g)})(x_{1j}^{(g)}-\hat{\mu}_1^{(g)}),\\
    s_{22} &= \sum_{g=1}^G\sum_{j=1}^{m_g}(x^{(g)}_{2j}-\hat{\mu}_2^{(g)})^2.
\end{align}

 Then, the resulting estimators obtained by maximizing $L$ w.r.t $\mu_1^{(g)}, \sigma_{11},\mu_2^{(g)}$, $\sigma_{22},\sigma_{12}$ are:
\begin{align*}
	\hat{\mu}_1^{(g)} & = \frac{1}{n_g}\sum_ {j=1}^{n_g}x_{1j}^{(g)},\;\;
	\hat{\mu}_2^{(g)} = \frac{\sum_ {j=1}^{m_g}x_{2j}^{(g)}+ \sum_ {j=n_g+1}^{l_g}x_{2j}^{(g)}}{m_g+l_g-n_g},	 \\
	\hat{\sigma}_{11} & = \frac{\sum_{g=1}^G\sum_ {j=1}^{n_g}(x_{1j}^{(g)}-\hat{\mu}_1^{(g)})^2}{\sum_{g=1}^G n_g},\\
\hat{\sigma}_{22} & = \frac{\sum_{g=1}^G[\sum_ {j=1}^{m_g}(x_{2j}^{(g)}-\hat{\mu}_2^{(g)})^2+ \sum_ {j=n_g+1}^{l_g}(x_{2j}^{(g)}-\hat{\mu}_2^{(g)})^2]}{\sum_{g=1}^G(m_g+l_g-n_g)}, 		
\end{align*}
 
$\hat{\sigma} _{12}$, where $\hat{\sigma} _{12}$ is the maximizer of:
	\begin{equation}\label{etacov}
	\eta(\sigma_{12}) = C-\frac{A}{2}\log \left(\sigma_{22}-\frac{\sigma_{12}^2}{\sigma_{11}}\right)
- \frac{s_{22} - 2\frac{\sigma_{12}}{\sigma_{11}} s_{12}+\frac{\sigma_{12}^2}{\sigma _{11}^2}s_{11}}  {2\left(\sigma_{22}-\frac{\sigma_{12}^2}{\sigma_{11}}\right)}.
	\end{equation}
\end{theorem}

This means the estimate for $\mu_1^{(g)}, \mu_2^{(g)},\sigma_{11}, \sigma_{22}$ are the corresponding sample means and uncorrected sample variance of the first/second feature after dropping the missing entries. We also have the following theorem regarding $\hat{\sigma}_{12}$,

\begin{theorem}\label{unique-multiple}
	For Equation (\ref{etacov}), solving
	\begin{equation}\label{der02}
		\frac{d\eta}{d\sigma_{12}} = 0
	\end{equation}
	can be reduced to finding the roots of the following third-degree polynomial 
	\begin{equation}\label{eqExist}
	s_{12} \sigma_{11} \sigma_{22} + {\left( \sigma_{11} \sigma_{22} A -  s_{22}\sigma_{11} - s_{11}\sigma_{22} \right)}\sigma_{12}  + s_{12}\sigma_{12}^2- A\sigma_{12}^3,
    \end{equation}
	which has at least one real root. Moreover, the global maximum is a real solution to that equation, provided that
	\begin{equation}
	-\frac{s_{22}\sigma_{11}+s_{11}\sigma_{22}}{2\sqrt{\sigma_{11}\sigma_{22}}} \neq
	s_{12} \neq \frac{s_{22}\sigma_{11}+s_{11}\sigma_{22}}{2\sqrt{\sigma_{11}\sigma_{22}}}. 
\end{equation}
\end{theorem}

The proof of this theorem can be given in \ref{existence}. In addition, the proof of Theorem \ref{bivariatemle} is provided in \ref{proof2mle}. It makes use of the following lemma, whose proof is available in \ref{sec:appendix_A}.

\begin{lemma}\label{lem1}
	Suppose that we have a data set that comes from $G$ classes, where the observations from the $g^{th}$ class follow multivariate normal distribution with mean $\tau^{{(g)}}$ and covariance matrix $\Delta$. Let the observations of the $g^{th}$ class be $\boldsymbol{u}^{(g)}_1,\boldsymbol{u}^{(g)}_2,...,\boldsymbol{u}^{(g)}_{n_g}$. Then, the MLEs of $\tau^{{(g)}}$ is the sample class mean $\bar{\mathbf{u} }^{(g)}$, and the MLE of $\Delta$ is 
	\begin{equation}
		\hat{\Delta} = \frac{1}{\sum_{g=1}^G n_g} \sum_{g=1}^G \sum_{j=1}^{n_g}
		(\boldsymbol{u} _j^{(g)}-\bar{\boldsymbol{u}}^{(g)})
		(\boldsymbol{u} _j^{(g)}-\bar{\boldsymbol{u}}^{(g)})'.
	\end{equation}
\end{lemma}

Based on Theorem \ref{bivariatemle}, we have the following algorithm.

\begin{Algorithm}{(DPER algorithm for multiple -class randomly missing data)} 

\textit{Input:} a data set of $G$ classes, $p$ features. 

\textit{Ouput:} $\hat{\boldsymbol{\mu}}^{(g)}, \hat{\boldsymbol{\Sigma}}=(\sigma_{ij})_{i,j=1}^p$

	\begin{enumerate}		
		
		\item Estimate $\boldsymbol{\mu}^{(g)}$: $\hat{\mu}_i^{(g)}$ is the mean of all the available entries in the $i^{th}$ feature. 
		
		\item Estimate the diagonal elements of $\boldsymbol{\Sigma}$: $\hat{\sigma}_{ii}$ is the uncorrected sample variance of all the available entries in the $i^{th}$ feature.  
		
		\item  For $1\le i\le p$:
		
		\;\;\; For $1\le j\le i$:
		
		\;\;\;\;\;\; Compute $\hat{\sigma}_{ij} = \hat{\sigma}_{ji}$ based on Equation (\ref{der02}). If there are two solutions maximizing the function, choose the one that is closest to the estimate based on  the case deletion method.
		
	\end{enumerate}
\end{Algorithm}

\begin{Remarks}
	We have the following remarks about our estimates:
	\begin{itemize}
		\item The computation is done on each pair of features. Therefore, we only need the assumption of bivariate normality on each pair of features instead of multivariate normality on each observation as in EPEM \cite{NGUYEN20211}. 
		
		\item The computation is done separately for each $\sigma_{ij}$.  As a result, our method is also applicable to the problems where the number of features is much higher than the sample size.
		
		\item Since maximum likelihood estimates are asymptotic (see \cite{casella2002statistical}),  we can ensure that the resulting estimates of the our algorithm are asymptotic, i.e.,
		\begin{equation}
		\hat{\boldsymbol {\mu } }\overset{p}{\rightarrow} \boldsymbol {\mu }, \text{and} \;\;\;	\hat{\boldsymbol {\Sigma } }\overset{p}{\rightarrow} \boldsymbol {\Sigma }, 
		\end{equation}
		where $\overset{p}{\rightarrow}$ denotes convergence in probability for each element of the matrix $\hat{\boldsymbol {\mu }}$ or $\hat{\boldsymbol {\Sigma }}$. 
		
	\end{itemize}
\end{Remarks}

\section{Experiments}\label{exper}\FloatBarrier
This section describes all data sets, the implementation of the proposed techniques, and experimental results in detail. 

\subsection{Settings}
To illustrate the efficiency of our algorithm, we compare the results of the proposed method the following methods: SOFT-IMPUTE \citep{mazumder2010spectral}, by Nuclear Norm Minimization \citep{candes2009exact}, MissForest \citep{stekhoven2012missforest}
and Multiple Imputation by Chained Equation (MICE) \citep{buuren2010mice}.  
We implement these methods via packages \textit{fancyimpute}\footnote{\url{https://pypi.org/project/fancyimpute/}}, \textit{impyute}\footnote{\url{https://pypi.org/project/impyute/}}, \textit{missingpy}\footnote{\url{https://pypi.org/project/missingpy/}}, and \textit{scikit-learn}\footnote{\url{https://scikit-learn.org/}}.

When using iterative methods for small data sets, we use the default options in the packages \textit{fancyimpute, sklearn} and \textit{impyute}. Specifically, the maximum number of iterations is 100 for SOFT-IMPUTE, MICE, Nuclear Norm Minimization, and 10 for MissForest
. For big data sets (MNIST \& Fashion MNIST), the maximum number of iterations for MICE and Nuclear Norm Minimization are reduced to 10 instead due to expensive memory and time requirements. 

The experiments on small data sets are ran directly on \textit{Google Colaboratory}\footnote{\url{https://colab.research.google.com/}}. Meanwhile, the experiments on large data sets are run on a Linux machine with 130 GB RAM, 40 processing units. We terminate an experiment if no result is produced after five hours of running or when having out-of-memory issues, and denote this as NA in the result tables (Tables \ref{table2} and \ref{table3}).

\subsection{Datasets}

\begin{table}
	\caption{Descriptions of data sets used in the experiments}
	\centering
	\label{tab1}
	\scalebox{0.82}{
		\begin{tabular}{|c|c|c|c|}
			\hline
			\textbf{data set}  &  \textbf{\# Classes}  & \textbf{ \# Features}  & \textbf{\# Samples} \\\cline{1-4}
			{Iris}& 3 &  4 &  150  \\  \hline
			{Wine}& 3 & 13 & 178 \\        \hline
			{Seeds} & 3 & 7 & 210 \\ \hline
			{Inosphere} & 2 & 34 {($32^*$)} & 351 \\
			\hline 
			{{MNIST}} & 10 &  {$784(649^*)$} & 70000\\
			\hline {{Fashion MNIST}} & 10 & $784$ & 70000\\
			\hline
	\end{tabular}}
\end{table}
We perform the experiments on the following data sets from the Machine Learning Database Repository at the University of California, Irvine \citep{Dua:2019}: Iris, Wine, Seeds, and Ionosphere. Besides, we do experiments on two large data sets: Fashion MNIST \cite{xiao2017fashion} and MNIST\cite{lecun1998mnist}. Fashion MNIST is a large data set of clothing images, consisting of 60000 training images and 10000 testing images of size $28\times 28$, and MNIST is a data set of handwritten digits, also having 60000 training images and 10000 testing images of size $28\times 28$. Each of these two data sets has ten labels. 

Table \ref{tab1} shows a summary of all data sets used in the experiment. For each data set, we normalize by removing the mean and scaling to unit variance all features. We delete the first column for Ionosphere, where the values are the same within one of the classes, and the second column, where all entries are 0. Lastly, for the MNIST, we delete 135 columns with no more than ten nonzero values (Note that we have ten classes. If one class has a column of all zeros, then the sample covariance matrix for that class is a degenerate matrix.)


\subsection{Evaluation Metrics}

For evaluation, we use the following metric \cite{NGUYEN20211}, which is the sum of the average difference of each entry in the mean/common covariance matrix.
\begin{equation}
r = \frac{||\boldsymbol {\mu } -\boldsymbol {\hat{\mu}}||_F }{n_{\boldsymbol {\mu } }}
+\frac{||\boldsymbol {\Sigma
	} -\boldsymbol {\hat{\Sigma}}||_F }{n_{\boldsymbol {\Sigma } }},
\end{equation}
where $||.||$ is the Frobenius norm; $\hat{\boldsymbol {\mu } }, \hat{\boldsymbol {\Sigma } }$ are estimated values derived from DPER for $\boldsymbol {\mu } , \boldsymbol {\Sigma } $ respectively; ${n_{\boldsymbol {\mu } }}$,  ${n_{\boldsymbol {\Sigma } }}$ are the corresponding number of entries in $\boldsymbol {\mu }, \boldsymbol {\Sigma }$ and the ground truth $\boldsymbol {\mu }, \boldsymbol {\Sigma }$ are calculated from the full data without missing values (because we do not know the true value of $\boldsymbol {\mu }, \boldsymbol {\Sigma }$). Note that unlike the mean square error, this evaluation measure takes into account the number of entries in the means and the covariance matrix instead.

\begin{table}
\caption{Parameters estimation errors without equal covariance matrix assumption}\label{table2}
\centering
\scalebox{0.82}{
\begin{tabular}{|c|c|c|c|c|c|c|c|}
\hline
 \textbf{data set} & \textbf{\makecell{Missing\\Rate(\%)}} & \textbf{DPER} & \textbf{MICE} & \textbf{\makecell{SOFT-\\IMPUTE}}  & \textbf{\makecell{Nuclear\\Norm}} & \textbf{MissForest} \\\hline
 Seeds & \makecell{20\%\\35\%\\50\%\\65\%\\80\%}  & \makecell{{\bf0.004}\\{\bf0.008}\\{\bf0.007}\\{\bf0.009}\\{\bf0.009}} &
 \makecell{{\bf0.004}\\0.009\\0.01\\0.017\\0.025} & \makecell{0.007\\0.015\\0.016\\0.025\\0.028} & \makecell{0.522\\0.517\\0.459\\0.489\\0.486} & \makecell{0.52\\0.543\\0.504\\0.607\\0.647} \\ \hline
 Iris & \makecell{20\%\\35\%\\50\%\\65\%\\80\%} & \makecell{{\bf0.007}\\{\bf0.009}\\{\bf0.006}\\{\bf0.013}\\{\bf0.024}}  &
 \makecell{0.009\\0.015\\0.017\\0.033\\0.035} & \makecell{0.013\\0.025\\0.026\\0.049\\0.049} & \makecell{0.693\\0.665\\0.597\\0.672\\0.616} & \makecell{0.71\\0.857\\0.812\\0.998\\0.951} \\ \hline
 Wine & \makecell{20\%\\35\%\\50\%\\65\%\\80\%} & \makecell{{\bf0.005}\\{\bf0.009}\\{\bf0.008}\\{\bf0.012}\\{\bf0.013}}  &
 \makecell{{\bf0.005}\\0.01\\0.011\\0.014\\0.018} & \makecell{0.006\\0.012\\0.016\\0.023\\0.025} & \makecell{0.27\\0.256\\0.239\\0.229\\0.215} & \makecell{0.288\\0.277\\0.276\\0.304\\0.286} \\ \hline
 Ionosphere & \makecell{20\%\\35\%\\50\%\\65\%\\80\%}  & \makecell{{\bf0.003}\\{\bf0.004}\\0.006\\0.007\\{\bf0.008}}  &
 \makecell{0.004\\0.005\\{\bf0.005}\\{\bf0.006}\\0.01} & \makecell{{\bf0.003}\\0.005\\{\bf0.005}\\0.007\\0.009} &
 \makecell{0.67\\0.615\\0.57\\0.531\\0.511} & \makecell{0.717\\0.704\\0.693\\0.688\\0.682} \\ \hline
 MNIST & \makecell{20\%\\35\%\\50\%\\65\%\\80\%}  &
 \makecell{{\bf0.4e-4}\\{\bf0.5e-4}\\{\bf0.7e-4}\\{\bf0.8e-4}\\{\bf0.9e-4}}& \makecell{NA\\NA\\NA\\NA\\NA} &
  \makecell{0.5e-4\\0.9e-4\\1.4e-4\\2e-4\\2.5e-4} &
 \makecell{NA\\NA\\NA\\NA\\NA} &
 \makecell{NA\\NA\\NA\\NA\\NA} \\ \hline
 Fashion MNIST & \makecell{20\%\\35\%\\50\%\\65\%\\80\%}  &
 \makecell{{\bf0.3e-4}\\{\bf0.4e-4}\\{\bf0.5e-4}\\{\bf0.6e-4}\\{\bf0.8e-4}}& \makecell{NA\\NA\\NA\\NA\\NA} &
 \makecell{0.4e-4\\0.7e-4\\0.9e-4\\1.3e-4\\1.6e-4} &
 \makecell{NA\\NA\\NA\\NA\\NA} &
 \makecell{NA\\NA\\NA\\NA\\NA} \\ \hline
\end{tabular}}
\end{table}
\begin{table}
\caption{Parameters estimation errors under the equal covariance matrix assumption}\label{table3}
\centering
\scalebox{0.82}{
\begin{tabular}{|c|c|c|c|c|c|c|c|} 
\hline
 \textbf{data set} & \textbf{\makecell{Missing\\Rate(\%)}} & \textbf{DPER} & \textbf{MICE} & \textbf{\makecell{SOFT-\\IMPUTE}}  & \textbf{\makecell{Nuclear\\Norm}} & \textbf{MissForest} \\\hline
 Seeds & \makecell{20\%\\35\%\\50\%\\65\%\\80\%}  & \makecell{0.004\\0.007\\{\bf0.009}\\{\bf0.008}\\{\bf0.014}}  &
 \makecell{{\bf0.003}\\{\bf0.006}\\0.013\\0.014\\0.034} & \makecell{0.005\\0.01\\0.02\\0.024\\0.033} & \makecell{0.005\\0.01\\0.019\\0.023\\0.025} & \makecell{0.004\\0.008\\0.011\\0.016\\0.019} \\ \hline
 Iris & \makecell{20\%\\35\%\\50\%\\65\%\\80\%} & \makecell{{\bf0.008}\\{\bf0.01}\\{\bf0.012}\\{\bf0.01}\\{\bf0.016}}  &
 \makecell{{\bf0.008}\\0.018\\0.029\\0.03\\0.049} & \makecell{0.014\\0.029\\0.043\\0.045\\0.069} & \makecell{0.014\\0.028\\0.04\\0.042\\0.061} & \makecell{0.01\\0.017\\0.021\\0.037\\0.042} \\ \hline
 Wine & \makecell{20\%\\35\%\\50\%\\65\%\\80\%} & \makecell{{\bf0.005}\\{\bf0.006}\\{\bf0.009}\\{\bf0.009}\\{\bf0.012}}  &
 \makecell{0.006\\0.009\\0.01\\0.014\\0.021} & \makecell{0.007\\0.011\\0.016\\0.021\\0.025} & \makecell{0.007\\0.011\\0.015\\0.02\\0.023} & \makecell{{\bf0.005}\\0.007\\0.01\\0.013\\0.017} \\ \hline
 Ionosphere & \makecell{20\%\\35\%\\50\%\\65\%\\80\%}  & \makecell{0.004\\0.005\\0.006\\0.008\\{\bf0.008}}  & \makecell{{\bf0.003}\\{\bf0.004}\\0.006\\{\bf0.007}\\0.01} &
 \makecell{{\bf0.003}\\{\bf0.004}\\0.006\\0.008\\0.009} & \makecell{{\bf0.003}\\{\bf0.004}\\0.006\\{\bf0.007}\\0.009} &
 \makecell{{\bf0.003}\\{\bf0.004}\\{\bf0.005}\\{\bf0.007}\\{\bf0.008}} \\ \hline
 MNIST & \makecell{20\%\\35\%\\50\%\\65\%\\80\%}  &
 \makecell{0.4e-4\\0.5e-4\\{\bf0.6e-4}\\0.8e-4\\1e-4}&
 \makecell{NA\\NA\\NA\\NA\\NA} &
 \makecell{{\bf0.3e-4}\\{\bf0.4e-4}\\{\bf0.6e-4}\\{\bf0.7e-4}\\{\bf0.9e-4}} &
 \makecell{NA\\NA\\NA\\NA\\NA} &
 \makecell{NA\\NA\\NA\\NA\\NA} \\ \hline
 Fashion MNIST & \makecell{20\%\\35\%\\50\%\\65\%\\80\%}  &
 \makecell{{\bf0.3e-4}\\{\bf0.4e-4}\\{\bf0.5e-4}\\{\bf0.6e-4}\\{\bf0.7e-4}}& \makecell{NA\\NA\\NA\\NA\\NA} & \makecell{0.4e-4\\0.7e-4\\1e-4\\1.3e-4\\1.7e-4} &
 \makecell{NA\\NA\\NA\\NA\\NA} &
 \makecell{NA\\NA\\NA\\NA\\NA} \\ \hline
\end{tabular}}
\end{table}

\subsection{Results and Discussion}
Noticeably, we do not conduct any experiment about single class randomly missing data. The reason is that the multiple-class problem assuming equal covariance matrices problem boils down to one-class sub-problems as explained in Section \ref{sec: without}. 

Table~\ref{table2} shows that without the assumption of equal covariance matrices, DPER gives the best result on every data set except for the Ionosphere. Interestingly, there is little distinction between DPER with MICE and SOFT-IMPUTE at lower missing rates, but the higher the missing rate, the better DPER performs against its peers. Considering the Seeds data set, for example, at $20\%$ missing rate, DPER gives the same best $0.004$ error rate as MICE, but at $80\%$ missing rate, the $0.009$ error rate of DPER is less than half of the others.

From Table~\ref{table3}, we see that under the equal covariance assumption, DPER gives the best results on all of the missing rates on every data sets except for Ionosphere and MNIST. Still, for Ionosphere and MNIST, DPER's performance is competitive against other approaches. When it does not perform best, the best method can only improve slightly than DPER (the maximum difference is $0.001$ for Ionosphere and $0.1e-4$ for MNIST). Again, there is little distinction between DPER with other approaches at lower missing rates, but the higher the missing rate, the better DPER performs against its peers. For instance, at $20\%$ missing rate in the Iris data set, the best error rate $0.008$ of DPER is equal to MICE, but at $80\%$ missing rate, the $0.016$ error rate of DPER is less than half of the rest.

It is worth noting whether to assume the covariance matrices are equal or not depends on the application and the nature of data sets. Demonstratively, from Tables~\ref{table2} and ~\ref{table3}, one can also see how this assumption affects the algorithms. First, the assumption seems to benefit DPER at high missing rates ($65\%, 80\%$) on small data sets (Seeds, Iris). For example, at $80\%$ of missing values on Iris, there is a $1/3$ reduction in the $0.024$ error rate of DPER with the equal covariance assumption. Yet, improvement may not happen in all small data sets with lower missing rates. For the Wine dataset, at $50\%$, the experiment with the assumption gives the error rate about $0.009$. Meanwhile, it is approximately $0.008$ without this assumption. There is no clear pattern on the effects of the equal covariance assumption for MICE and SOFT-IMPUTE datasets. 

Meanwhile, Nuclear Norm and MissForest datasets have a significantly better performance with the assumption. As an indication, with $20\%$ missing rate in the Ionosphere data set, the $0.717$ error rate of MissForest reduces over $99\%$ to $0.005$ in Table~\ref{table3}. The result suggests that for smaller data sets, the equal covariance matrix assumption benefits Nuclear Norm and MissForest significantly. It may be due to the number of samples used for the covariance estimation is lower without the assumption.

The parameter estimation on large data sets such as MNIST and Fashion MNIST, DPER gives small error rates. Since the populational means and covariance matrix are unknown, one can evaluate the estimation on the complete data sets (i.e., the data sets before the missing data simulation). It implies that with DPER, we can approximate means and covariance matrices with almost identical results with the complete data set. This result is reasonable since the Multivariate Central Limit Theorem suggests the maximum likelihood parameter approximation gets better with the increment of sample size, and these data sets are large.
Remarkable, only the results for DPER and SOFT-IMPUTE are available for these two data sets. Among the available methods, DPER is the fastest one with minimum run-time for parameter estimation on MNIST at $69.54$s, while SOFT-IMPUTE is the second fastest ($165.15$s), respectively. For other approaches, the corresponding outcomes are not available due to memory-bound or running-time bound. Therefore, we can deduce that DPER and SOFT-IMPUTE are more suitable than MICE, Nuclear Norm, and MissForest for large data sets.

\section{Conclusion}\label{sec:conclusion}

In this paper, we have derived formulas and provided algorithms to compute the mean and covariance matrices from the data suitable for both small and large data sets, hence providing an approximation to the data's underlying distribution(s). We illustrate the power of our methods by experiments, which show that while iterative methods may not handle the missing data problem for big data sets, our approach can still find the parameter rapidly without requiring extensive memory and computation time. As the estimation is done for each pair of features, the proposed method is also applicable to problems where the number of features is much higher than the sample size.

Yet, our methods still have several drawbacks and room for improvement. 
First, when the distribution of the data is highly skewed \cite{mardia1970measures}, the bivariate normal distribution might not be a good approximator. Therefore, when the pair of features is highly skewed, other methods rather than DPER may be utilized for that pair of features.  

Second, the computation is done separately for each $\sigma_{ij}$ that only requires the data of the $i$th and $j$th features. It also means we have not used the extra information from the data set's other features. Also, separate computation for each of $\sigma_{ij}$ means we could parallelize the algorithm even further. These will be the topics for our future research. We also publish all codes related to the experiments in one Github repository\footnote{\url{https://github.com/thunguyen177/DPER}}.

\section*{Acknowledgments}

We want to thank the University of Science, Vietnam National University in Ho Chi Minh City, and  AISIA Research Lab in Vietnam for supporting us throughout this paper. This research is funded by Vietnam National University Ho Chi Minh City (VNU-HCM) under grant number C2021-18-03.

\bibliography{ref}

\newpage
\appendix
\section{Proof of Theorem \ref{lem1}}
\label{sec:appendix_A}

\textit{Proof.} The log-likelihood of the data is
\begin{align}
	l & = \sum_{g=1}^G \sum_{j=1}^{n_g} \log f(\boldsymbol{u}_j^{(g)})\\
	& = C - \frac{1}{2}\sum_{g=1}^Gn_g\log|\Delta| - \frac{1}{2}\sum_{g=1}^G \sum_{j=1}^{n_g} \delta_j^{(g)},
\end{align}
where 
\begin{equation}
	\delta_j^{(g)}=	(\boldsymbol{u}_j^{(g)}-\boldsymbol {\tau}^{(g)})'\Delta^{-1}(\boldsymbol{u}_j^{(g)}-\boldsymbol {\tau}^{(g)}).
\end{equation}
We have
\begin{align*}
	\frac{\partial \delta_j^{(g)}}{\partial \boldsymbol {\tau} ^{(g)}} & = \frac{\partial \delta_j^{(g)}}{\partial(\boldsymbol{u}_j^{(g)}- \boldsymbol {\tau} ^{(g)})}
	\frac{\partial(\boldsymbol{u}_j^{(g)}- \boldsymbol {\tau} ^{(g)})}{\partial \boldsymbol {\tau} ^{(g)}}\\
	&= \Delta^{-1}(\boldsymbol{u}_j^{(g)}- \boldsymbol {\tau} ^{(g)}).
\end{align*}
Hence,
\begin{align*}
	\frac{\partial l}{\partial \boldsymbol {\mu } _j^{(g)}} & \Leftrightarrow  \sum_{g=1}^G \sum_{j=1}^{n_g}(\boldsymbol{u}_j^{(g)}- \boldsymbol {\tau} ^{(g)})=0\\
	& \Leftrightarrow \hat{ \boldsymbol {\tau}} ^{(g)} = \bar{\mathbf{u} }^{(g)}.
\end{align*}

To find $\hat{\Delta}$, note that
\begin{equation}
	\frac{\partial \log |\Delta|}{\partial \Delta^{-1}}=-\frac{\partial \log |\Delta^{-1}|}{\partial \Delta^{-1}}=-\Delta.
\end{equation}
Next,
\begin{equation}
	\delta_j^{(g)}=	tr(\Delta^{-1}(\boldsymbol{u}_j^{(g)}-\boldsymbol {\tau}^{(g)})(\boldsymbol{u}_j^{(g)}-\boldsymbol {\tau}^{(g)})').
\end{equation}
Therefore,
\begin{equation}
	\frac{\partial \delta_j^{(g)}}{\partial \Delta^{-1}} = (\boldsymbol{u}_j^{(g)}-\boldsymbol {\tau}^{(g)})(\boldsymbol{u}_j^{(g)}-\boldsymbol {\tau}^{(g)})'.
\end{equation}
This implies
\begin{align}
	\frac{\partial l}{\partial \Delta^{-1}}=0 & \Leftrightarrow 
	\sum_{g=1}^G \sum_{j=1}^{n_g}
	(\boldsymbol{u}_j^{(g)}-\boldsymbol {\tau}^{(g)})(\boldsymbol{u}_j^{(g)}-\boldsymbol {\tau}^{(g)})'=0\\
	& \Rightarrow \hat{\Delta} = \frac{1}{\sum_{g=1}^G n_g}\sum_{g=1}^G \sum_{j=1}^{n_g}(\boldsymbol{u}_j^{(g)}-\boldsymbol {\tau}^{(g)})(\boldsymbol{u}_j^{(g)}-\boldsymbol {\tau}^{(g)})'.
\end{align}
  
\section{Proof of Theorem \ref{bivariatemle}}\label{proof2mle}
%
Let $\boldsymbol{u}_j^{(g)}$ be the $j^{th}$ column of $\boldsymbol{x} ^{(g)}$ and use $f(\boldsymbol{z})$ to denote the joint density of the available entries in a random vector $\boldsymbol{z}$.  

\textbf{\textit{Maximizing the likelihood w.r.t $\hat{\mu}_1^{(g)}, \sigma_{11}$}}

It is worth noting that for $j=1,2,...,m_g$,
\begin{equation}
	f(\boldsymbol{u} _j^{(g)}) =f(\boldsymbol{x}_{1j}^{(g)},\boldsymbol{x}_{2j}^{(g)}) 
	= f(\boldsymbol{x}_{1j}^{(g)})f(\boldsymbol{x}_{2j}^{(g)}|\boldsymbol{x}_{1j}^{(g)}).
\end{equation} 
Hence, the likelihood function is
\begin{equation}
\small
	L = \prod_{g=1}^G\left[\left(\prod_{j=1}^{m_g}f(\boldsymbol{x}_{1j}^{(g)})f(\boldsymbol{x}_{2j}^{(g)}|\boldsymbol{x}_{1j}^{(g)})\right)
	\left(\prod_{j=m_g+1}^{n_g}f(\boldsymbol{x}_{1j}^{(g)})\right)
	\left(\prod_{j=n_g+1}^{l_g}f(\boldsymbol{x}_{2j}^{(g)})\right)\right], 
\end{equation}
which can be rewritten as
\begin{equation}
\small
	L = \prod_{g=1}^G\left[\left(\prod_{j=1}^{m_g}f(\boldsymbol{x}_{2j}^{(g)}|\boldsymbol{x}_{1j}^{(g)})\right)
	\left(\prod_{j=1}^{n_g}f(\boldsymbol{x}_{1j}^{(g)})\right)
	\left(\prod_{j=n_g+1}^{l_g}f(\boldsymbol{x}_{2j}^{(g)})\right)\right].
\end{equation}
It turns out that maximizing $L$ w.r.t $ {\mu }_1^{(g)},  {\sigma }_{11}$ is equivalent to maximizing $\prod_{g=1}^G\prod_{j=1}^{n_g}f(\boldsymbol{x}_{1j}^{(g)})$ w.r.t   $ {\mu }_1^{(g)},  {\sigma }_{11}$. This problem is similar to the problem of finding the MLEs in Lemma \ref{lem1}. Therefore, by applying  Lemma  \ref{lem1}, the MLEs for  $ {\mu }_1^{(g)},  {\sigma }_{11}$ are
\begin{align}  
	\hat{\mu}_1^{(g)} & = \frac{1}{n_g}\sum_ {j=1}^{n_g}x_{1j}^{(g)},\\
	\hat{\sigma}_{11}^{(g)} & = \frac{\sum_ {j=1}^{n_g}\sum_{g=1}^G(x_{1j}^{(g)}-\hat{\mu}_1^{(g)})^2}{\sum_{g=1}^G n_g}.
\end{align}

\textbf{\textit{Maximizing the likelihood w.r.t $\hat{\mu}_2^{(g)}, \sigma_{22}$}}

For $j=1,2,...,m_g,$
\begin{equation}
	f(\boldsymbol{u} _j^{(g)}) =f(\boldsymbol{x}_{1j}^{(g)},\boldsymbol{x}_{2j}^{(g)}) 
	= f(\boldsymbol{x}_{2j}^{(g)})f(\boldsymbol{x}_{1j}^{(g)}|\boldsymbol{x}_{2j}^{(g)}).
\end{equation} 
Hence, the likelihood function is
\begin{equation}
	\small
	L = \prod_{g=1}^G\left[\left(\prod_{j=1}^{m_g}f(\boldsymbol{x}_{2j}^{(g)})f(\boldsymbol{x}_{1j}^{(g)}|\boldsymbol{x}_{2j}^{(g)})\right)
	\left(\prod_{j=m_g+1}^{n_g}f(\boldsymbol{x}_{1j}^{(g)})\right)
	\left(\prod_{j=n_g+1}^{l_g}f(\boldsymbol{x}_{2j}^{(g)})\right)\right].
\end{equation}
Consequently, maximizing $L$ w.r.t $ {\mu }_2^{(g)},  {\sigma }_{22}$ is equivalent to maximizing 
\begin{equation}
	\prod_{g=1}^G\left[\left(\prod_{j=1}^{m_g}f(\boldsymbol{x}_{2j}^{(g)})\right)\left(\prod_{j=n_g+1}^{l_g}f(\boldsymbol{x}_{2j}^{(g)})\right)\right]
\end{equation}
w.r.t   $ {\mu }_2^{(g)},  {\sigma }_{22}$. This problem is similar to the problem of finding the MLEs in Lemma \ref{lem1}. Therefore, by applying  Lemma  \ref{lem1}, the MLEs for  $ {\mu }_2^{(g)},  {\sigma }_{22}$ are
\begin{align*}
	\hat{\mu}_2^{(g)} &= \frac{\sum_ {j=1}^{m_g}x_{2j}^{(g)}+ \sum_ {j=n_g+1}^{l_g}x_{2j}^{(g)}}{m_g+l_g-n_g},	 \\
	\hat{\sigma}_{22}^{(g)} & = \frac{\sum_{g=1}^G[\sum_ {j=1}^{m_g}(x_{2j}^{(g)}-\hat{\mu}_2^{(g)})^2+ \sum_ {j=n_g+1}^{l_g}(x_{2j}^{(g)}-\hat{\mu}_2^{(g)})^2]}{\sum_{g=1}^G(m_g+l_g-n_g)}.		
\end{align*}

\textbf{\textit{Maximizing the likelihood function with respect to $\sigma_{12}$.}}

Maximizing $\log L$ w.r.t $\sigma_{12}$ is equivalent to  maximizing
\begin{equation}
\eta = \sum_{g=1}^G\sum_{j=1}^{m_g}\log f(x_{2j}^{(g)}|x_{1j}^{(g)})
\end{equation}
w.r.t $\sigma _{12}$.

Note that $x_{2j}^{(g)}|x_{1j}^{(g)}$ follows normal distribution with mean
\begin{equation}
\hat{\mu}_2^{(g)} + \frac{\sigma_{12}}{\sigma _{11}}(x_{1j}^{(g)}-\hat{\mu}_1^{(g)}),
\end{equation}
and variance
\begin{equation}
    \sigma_{22}-\frac{\sigma_{12}^2}{\sigma_{11}}.
\end{equation}
Consequently,
\begin{equation}
\eta = C-\frac{1}{2}\sum_{g=1}^Gm_g\log \left(\sigma_{22}-\frac{\sigma_{12}^2}{\sigma_{11}}\right)
- \frac{1}{2} \left(\sum_{g=1}^G\sum_{j=1}^{m_g}\delta_{gj}^2\right)\left(\sigma_{22}-\frac{\sigma_{12}^2}{\sigma_{11}}\right)^{-1},
\end{equation}
where 
\begin{equation}
\delta_{gj} = x^{(g)}_{2j}-\hat{\mu}_2^{(g)} - \frac{\sigma_{12}}{\sigma _{11}}(x_{1j}^{(g)}-\hat{\mu}_1^{(g)}).
\end{equation}
One can see that
\begin{align}
\delta_{gj}^2 &= ((x^{(g)}_{2j}-\hat{\mu}_2^{(g)}) - \frac{\sigma_{12}}{\sigma _{11}}(x_{1j}^{(g)}-\hat{\mu}_1^{(g)}))^2\\
&= (x^{(g)}_{2j}-\hat{\mu}_2^{(g)})^2 - 2(x^{(g)}_{2j}-\hat{\mu}_2^{(g)})\frac{\sigma_{12}}{\sigma _{11}}(x_{1j}^{(g)}-\hat{\mu}_1^{(g)})
+ \frac{\sigma_{12}^2}{\sigma _{11}^2}(x_{1j}^{(g)}-\hat{\mu}_1^{(g)})^2.
\end{align}
As a result,
\begin{equation}
\sum_{g=1}^G\sum_{j=1}^{m_g}\delta_{gj}^2= s_{22} - 2\frac{\sigma_{12}}{\sigma_{11}} s_{12}+\frac{\sigma_{12}^2}{\sigma _{11}^2}s_{11}.
\end{equation}
Hence, $\hat{\sigma}_{12}$ is the maximizer of 
\begin{equation}
	\eta(\sigma_{12}) = C-\frac{A}{2}\log \left(\sigma_{22}-\frac{\sigma_{12}^2}{\sigma_{11}}\right)
- \frac{s_{22} - 2\frac{\sigma_{12}}{\sigma_{11}} s_{12}+\frac{\sigma_{12}^2}{\sigma _{11}^2}s_{11}}  {2\left(\sigma_{22}-\frac{\sigma_{12}^2}{\sigma_{11}}\right)}.
\end{equation}

\section{Proof of Theorem \ref{unique-multiple}} \label{existence}
In the following proof, when a real-valued function $h(x)$ has two one-sided limits $\lim_{x \to -\sqrt{a}^{+}} h(x), \lim_{x \to \sqrt{a}^{-}} h(x)$ that coincide, we simultaneously refer to them as
\begin{align*}
	\lim_{x^2 \to a^{-}} h(x) &= \lim_{x \to -\sqrt{a}^{+}} h(x) = \lim_{x \to \sqrt{a}^{-}} h(x).
\end{align*}
Provided that the left-handed limits are equal to their corresponding right-handed limits and handedness of the limits are preserved with our relevant calculation, a true result of one one-sided limit is also true for another. The use of the notation $\lim_{x^2 \to a^{-}} h(x)$ is then to avoid redundancy.

The proof attempts to show that the global maximum point of $\eta$ lies inside the domain, hence is a solution to $\frac {d\eta}{d\sigma_{12}}=0$. It should be reminded that in our problem, $\sigma_{12}\in (-\sqrt{\sigma_{11}\sigma_{22}},\sqrt{\sigma_{11}\sigma_{22}})$, $A \geq 0$, $\sigma_{11}>0$, $\sigma_{22}>0$ and \begin{equation}
	-\frac{s_{22}\sigma_{11}+s_{11}\sigma_{22}}{2\sqrt{\sigma_{11}\sigma_{22}}} \neq
	s_{12} \neq \frac{s_{22}\sigma_{11}+s_{11}\sigma_{22}}{2\sqrt{\sigma_{11}\sigma_{22}}}. \label{cond7}
\end{equation}

For $1\leq g \leq G$, $1 \leq j \leq m_g$, we define
\begin{align*}
	a_{g,j}&=\frac{\sigma_{12}}{\sigma_{11}}(x_{1j}^{(g)}-\hat{\mu}_1^{(g)}), \\
	b_{g,j} &= (x_{2j}^{(g)}-\hat{\mu}_2^{(g)}), \\
	q &= s_{22} - 2\frac{\sigma_{12}}{\sigma_{11}} s_{12}+\frac{\sigma_{12}^2}{\sigma _{11}^2}s_{11}, \\
	\phi(\sigma_{12}) &= -\frac{A}{2}\log \left(\sigma_{22}-\frac{\sigma_{12}^2}{\sigma_{11}}\right), \\
	\psi(\sigma_{12}) &= \frac{s_{22} - 2\frac{\sigma_{12}}{\sigma_{11}} s_{12}+\frac{\sigma_{12}^2}{\sigma _{11}^2}s_{11}}  {2\left(\sigma_{22}-\frac{\sigma_{12}^2}{\sigma_{11}}\right)}.
\end{align*} 
\textit{Proof.}

Note that $a_{g,j}^2+b_{g,j}^2 - 2a_{g,j}b_{g,j} \geq 0$ for every pair of $(g,j)$. By summing across all pair of indices, it follows that 
\begin{align*}
    q = \sum_{g=1}^{G}\sum_{j=1}^{m_g} (a_{g,j}-b_{g,j})^2 \geq 0
\end{align*}
From the condition~\ref{cond7},
\begin{align*}
	\sigma_{11}q(-\sqrt{\sigma_{11}\sigma_{22}}) = 
	s_{22}\sigma_{11} + s_{11} \sigma_{22} + 2 \, \sqrt{\sigma_{11} \sigma_{22}} s_{12} & > 0, \\
	\sigma_{11}q(\sqrt{\sigma_{11}\sigma_{22}}) = 
	s_{22}\sigma_{11} + s_{11} \sigma_{22} - 2 \, \sqrt{\sigma_{11} \sigma_{22}} s_{12} & > 0.
\end{align*}
As a consequence, 
\begin{align*}
	\lim_{\sigma_{12} \to -\sqrt{\sigma_{11} \sigma_{22}}^{+}} \psi(\sigma_{12}) &= \frac {q(-\sqrt{\sigma_{11} \sigma_{22}})} {0^+} = \infty,\\ 
	\lim_{\sigma_{12} \to \sqrt{\sigma_{11} \sigma_{22}}^{-}} \psi(\sigma_{12}) &= \frac {q(\sqrt{\sigma_{11} \sigma_{22}})} {0^+} = \infty.
\end{align*}
In our notation, that is $\lim_{\sigma_{12}^2 \to {\sigma_{11} \sigma_{22}}^{-}} \psi(\sigma_{12}) = \infty$. 

\textit{Case of $A>0$:} 

 It can be seen that 
 \begin{equation*}
     \lim_{\sigma_{12}^2 \to {\sigma_{11} \sigma_{22}}^{-}} \phi(\sigma_{12}) = \infty.
 \end{equation*}
 Moreover,
\begin{align*}
	&\lim_{\sigma_{12} \to -\sqrt{{\sigma_{11} \sigma_{22}}}^{+}} \frac{-\psi(\sigma_{12})}{\phi(\sigma_{12})} \\
	&=\lim_{\sigma_{12} \to -\sqrt{{\sigma_{11} \sigma_{22}}}^{+}} \frac{s_{22} \sigma_{11} \sigma_{12} - s_{12} \sigma_{12}^{2} - {\left(s_{12} \sigma_{11} - s_{11} \sigma_{12}\right)} \sigma_{22}}{A \sigma_{12}(\sigma_{12}^{2}-\sigma_{11}\sigma_{22})}\\
	&=\frac{-\sqrt{\sigma_{11}\sigma_{22}} \left ( 2 \, s_{12} \sqrt{\sigma_{11}\sigma_{22}} + s_{22}\sigma_{11}+s_{11}\sigma_{22} \right )}{0^{+}} = -\infty,\\
	&\lim_{\sigma_{12} \to \sqrt{{\sigma_{11} \sigma_{22}}}^{-}} \frac{-\psi(\sigma_{12})}{\phi(\sigma_{12})}\\
	&=\lim_{\sigma_{12} \to \sqrt{{\sigma_{11} \sigma_{22}}}^{-}} \frac{s_{22} \sigma_{11} \sigma_{12} - s_{12} \sigma_{12}^{2} - {\left(s_{12} \sigma_{11} - s_{11} \sigma_{12}\right)} \sigma_{22}}{A \sigma_{12}(\sigma_{12}^{2}-\sigma_{11}\sigma_{22})}\\
	&=\frac{\sqrt{\sigma_{11}\sigma_{22}} \left ( -2 \, s_{12} \sqrt{\sigma_{11}\sigma_{22}} + s_{22}\sigma_{11}+s_{11}\sigma_{22} \right )}{0^{-}} = -\infty.
\end{align*}
It follows that
\begin{align*}
	& \lim_{\sigma_{12}^2 \to {\sigma_{11} \sigma_{22}}^{-}} \eta(\sigma_{12})  \\
	&= C + \lim_{\sigma_{12}^2 \to {\sigma_{11} \sigma_{22}}^{-}} \phi(\sigma_{12})-\psi(\sigma_{12})\\
	&= C + \lim_{\sigma_{12}^2 \to {\sigma_{11} \sigma_{22}}^{-}}\left( 1+\frac{-\psi(\sigma_{12})}{\phi(\sigma_{12})} \right)
	\lim_{\sigma_{12}^2 \to {\sigma_{11} \sigma_{22}}^{-}}
	\phi(\sigma_{12})\\
	&= C + (-\infty)\infty = -\infty.
\end{align*}
It can be derived that
$$\eta'(\sigma_{12}) = \frac{s_{12} \sigma_{11} \sigma_{22} + {\left( \sigma_{11} \sigma_{22} A -  s_{22}\sigma_{11} - s_{11}\sigma_{22} \right)}\sigma_{12}  + s_{12}\sigma_{12}^2- A\sigma_{12}^3 } {{\left( \sigma_{12}^{2} - \sigma_{11} \sigma_{22}\right)}^2}.$$ Since $\lim_{\sigma_{12}^2 \to {\sigma_{11} \sigma_{22}}^{-}} \eta(\sigma_{12}) = -\infty$ and $\eta$ is twice differentiable over its domain, 
it has a global maximum point in the domain, which is a real root of
\begin{equation}\label{eqc.2}
	s_{12} \sigma_{11} \sigma_{22} + {\left( \sigma_{11} \sigma_{22} A -  s_{22}\sigma_{11} - s_{11}\sigma_{22} \right)}\sigma_{12}  + s_{12}\sigma_{12}^2- A\sigma_{12}^3.
\end{equation}
\textit{Case of $A=0$:} 

The function becomes $\eta = C - \psi$. Consequently, $\lim_{\sigma_{12}^2 \to {\sigma_{11} \sigma_{22}}^{-}} \eta(\sigma_{12}) = -\infty$. The derivative is now 
\begin{equation*}
	\eta' = \frac{s_{12} \sigma_{11} \sigma_{22} - {\left(s_{22}\sigma_{11} + s_{11}\sigma_{22} \right)}\sigma_{12}  + s_{12}\sigma_{12}^2} {{\left( \sigma_{12}^{2} - \sigma_{11} \sigma_{22}\right)}^2}.
\end{equation*}
With the same reasoning as the last paragraph, $\eta$ has a global maximum point in the domain that is a root of \ref{eqc.2}. 


\end{document}